\documentclass[11pt,a4paper]{article}
\usepackage[hyperref]{naaclhlt2018}
\usepackage{times}
\usepackage{latexsym}
\usepackage{graphicx}
\usepackage{url}
\usepackage{listings}

\aclfinalcopy

\newcommand{\ignore}[1]{}

\title{Strong Baselines for Simple Question Answering over\\
  Knowledge Graphs with and without Neural Networks}

\author{Salman Mohammed, Peng Shi, and Jimmy Lin \\
  David R. Cheriton School of Computer Science \\
  University of Waterloo \\
  {\tt \small smohammed1993@gmail.com, \{peng.shi,jimmylin\}@uwaterloo.ca}}

\date{}

\begin{document}
\maketitle
\begin{abstract}
We examine the problem of question answering over knowledge graphs,
focusing on simple questions that can be answered by the lookup of a
single fact. Adopting a straightforward decomposition of the problem
into entity detection, entity linking, relation prediction, and
evidence combination, we explore simple yet strong baselines.
On the popular \textsc{Simple\-Questions} dataset, we
find that basic LSTMs and GRUs plus a few heuristics yield
accuracies that approach the state of the art, and techniques that
do not use neural networks also perform reasonably well. These results
show that gains from sophisticated deep learning techniques
proposed in the literature are quite modest and that some previous models
exhibit unnecessary complexity.
\end{abstract}

\section{Introduction}

There has been significant recent interest in {\it simple}
question answering over knowledge graphs, where a natural language
question such as ``Where was Sasha Vujacic born?''\ can be answered
via the lookup of a simple fact---in this case, the ``place of birth''
property of the entity ``Sasha Vujacic''. Analysis of an existing
benchmark dataset~\cite{YaoXuchen_NAACL-HLT2015} and real-world user
questions~\cite{DaiZihang_etal_ACL2016,Ture_Jojic_EMNLP2017}
show that such questions cover a broad range of users' needs.

Most recent work on the simple QA task involves increasingly
complex neural network (NN) architectures that yield
progressively smaller gains over the previous state of the art (see
\S\ref{section:related} for more details). Lost in this push, we
argue, is an understanding of what exactly contributes to the
effectiveness of a particular NN architecture.
In many cases, the lack of rigorous ablation studies further
compounds difficulties in interpreting results and credit assignment.
To give two related examples:\ \citet{Gabor_etal_2017} reported that standard
LSTM architectures, when properly tuned, outperform some more recent
models; \citet{Vaswani_etal_2017} showed that the dominant approach to
sequence transduction using complex encoder--decoder networks with
attention mechanisms work just as well with the attention module only,
yielding networks that are far simpler and easier to train.

In line with an emerging thread of research that aims to improve
empirical rigor in our field by focusing on knowledge and
insights, as opposed to simply ``winning''~\cite{Sculley_etal_2018},
we take the approach of peeling away unnecessary complexity until we
arrive at the simplest model that works well.
On the \textsc{SimpleQuestions} dataset,
we find that baseline NN architectures plus simple heuristics
yield accuracies that approach the state of the art. Furthermore, we
show that a combination of simple techniques that do not involve neural networks can
still achieve reasonable accuracy. These results suggest that while
NNs do indeed contribute to meaningful advances on this
task, some models exhibit unnecessary complexity and that
the best models yield at most modest gains over strong baselines.

\section{Related Work}
\label{section:related}

The problem of question answering on knowledge graphs dates back
at least a decade, but the most relevant recent work in the NLP
community comes from \citet{Berant_etal_EMNLP2013}.
This thread of work focuses
on semantic parsing, where a question is mapped to its
logical form and then translated to a structured
query, cf.~\cite{Berant_Liang_ACL2014,Reddy_etal_TACL2014}.
However, the more recent \textsc{SimpleQuestions} dataset~\cite{Bordes_etal_2015}
has emerged as the de facto
benchmark for evaluating simple QA over knowledge graphs.

\hspace{-0.05cm}The original solution of \citet{Bordes_etal_2015} featured memory
networks, but over the past several years, researchers have applied
many NN architectures for tackling this
problem:\ \citet{Golub_He_EMNLP2016} proposed a character-level
attention-based encoder-decoder framework; \citet{DaiZihang_etal_ACL2016} 
proposed a conditional probabilistic framework using BiGRUs. 
\citet{Lukovnikov_etal_WWW2017} used a hierarchical word/character-level 
question encoder and trained a neural network in an end-to-end manner.
\citet{YinWengpeng_etal_2016} applied a
character-level CNN for entity linking and a separate word-level
CNN with attentive max-pooling for fact selection.
\citet{YuMo_etal_ACL2017} used a hierarchical residual BiLSTM
for relation detection, the results of which were combined
with entity linking output. These approaches can be characterized
as exploiting increasingly sophisticated modeling techniques
(e.g., attention, residual learning, etc.). 

In this push toward complexity, we do not believe that researchers
have adequately explored baselines, and thus it is unclear how much
various NN techniques actually help. To this end, our work
builds on \citet{Ture_Jojic_EMNLP2017}, who adopted a straightforward
problem decomposition with simple NN models to argue that
attention-based mechanisms don't really help. We take this one step further
and examine techniques that do not involve neural networks.
Establishing strong baselines allows us to objectively quantify the
contribution of various deep learning techniques.

\section{Approach}

We begin with minimal preprocessing on questions:\ downcasing and
tokenizing based on the
Penn TreeBank. As is common in the literature, we decompose the simple QA
problem into four tasks:\ entity detection, entity
linking, relation prediction, and evidence integration, detailed below.
All our code is available open source on
GitHub.\footnote{http://buboqa.io/}

\subsection{Entity Detection}

Given a question, the goal of entity detection is to identify the
entity being queried. This is naturally formulated as a
sequence labeling problem, where for each token, the task is to assign
one of two tags, either \textsc{Entity} or \textsc{Not\-Entity}.

\smallskip \noindent \textbf{Recurrent Neural Networks (RNNs):} The
most obvious NN model for this task is to use RNNs;
we examined both bi-directional LSTM and GRU variants over an
input matrix comprised of word embeddings from the input
question. Following standard practice, the representation of each
token is a concatenation of the hidden states from the forward and
backward passes. This representation is then passed through a linear
layer, followed by batch normalization, ReLU activation, dropout, and
a final layer that maps into the tag space. Note that since we're
examining baselines, we do {\it not} layer a CRF on top 
of the BiLSTM~\cite{Lample_etal_NAACL-HLT2016,MaXuezhe_Hovy_ACL2016}.

\smallskip \noindent \textbf{Conditional Random Fields (CRFs):} Prior
to the advent of neural techniques, CRFs represented the state of the
art in sequence labeling, and therefore it makes sense to explore how
well this method works. We specifically adopt the approach of
\citet{Finkel_etal_ACL2005}, who used features such as word
positions, POS tags, character n-grams, etc.

\subsection{Entity Linking}   

The output of entity detection is a sequence of tokens representing a
candidate entity. This still needs to be linked to an actual node in
the knowledge graph. In Freebase, each node is denoted by a Machine Identifier, or MID. Our
formulation treats this problem as fuzzy string matching and does not
use neural networks.

For all the entities in the knowledge graph (Freebase), we pre-built an inverted
index over $n$-grams $ n \in \{1,2,3\}$ in an entity's name. At
linking time, we generate all corresponding $n$-grams from the
candidate entity and look them up in the inverted index for all
matches. Candidate entity MIDs are retrieved from the index and
appended to a list, and an early termination heuristic similar to
\citet{Ture_Jojic_EMNLP2017} is applied. We start with $n=3$ and if we find
an exact match for an entity, we do not further consider lower-order
$n$-grams, backing off otherwise. Once all candidate entities
have been gathered, they are then
ranked by Levenshtein Distance to the MID's canonical label.

\subsection{Relation Prediction}

The goal of relation prediction is to identify the relation being
queried. We view this as classification over the entire
question.

\smallskip \noindent \textbf{RNNs:} Similar to entity detection, we
explored BiLSTM and BiGRU variants. Since relation
prediction is over the entire question, we base the
classification decision only on the hidden states (forward and
backward passes) of the final token, but otherwise the model
architecture is the same as for entity detection.

\smallskip \noindent \textbf{Convolutional Neural Networks (CNNs):} 
Another natural model is to use CNNs, which
have been shown to perform well for sentence classification. We adopt the
model of \citet{Kim_EMNLP2014}, albeit slightly simplified in that we
use a single static channel instead of multiple channels.
Feature maps of widths two to
four are applied over the input matrix comprised of input tokens
transformed into word embeddings, followed by max pooling, a
fully-connected layer and softmax to output the final prediction.
Note this is a ``vanilla'' CNN without any attention
mechanism.

\smallskip \noindent \textbf{Logistic Regression (LR):}\ Before
the advent of neural networks, the most obvious solution
to sentence classification would be to apply logistic regression.
We experimented with two feature sets over the questions:\ (1) tf-idf on
unigrams and bigrams and (2) word embeddings +
relation words. In~(2), we averaged the word embeddings of each token in the question,
and to that vector, we concatenated the one-hot vector
comprised of the top 300 most frequent terms from the names of the relations
(e.g., people/person/place\_of\_birth),
which serve as the dimensions of the one-hot vector.
The rationale behind this hybrid
representation is to combine the advantages of word embeddings in
capturing semantic similarity with the ability of one-hot vectors to
clearly discriminate strong ``cue'' tokens in the relation names.

\subsection{Evidence Integration} 

Given the top $m$ entities and $r$ relations from
the previous components, the final task is to integrate evidence to
arrive at a single (entity, relation) prediction. We begin by
generating $m \times r$ (entity, relation) tuples whose scores are the
product of their component scores. Since both entity detection/linking
and relation prediction are performed independently, many
combinations are meaningless (e.g., no such relation exists
for an entity in the knowledge graph); these are pruned.

After pruning, we observe many scoring ties, which arise from nodes in
the knowledge graph that share the exact same label, e.g., all persons
with the name ``Adam Smith''. We break ties by
favoring more ``popular'' entities, using the number of incoming edges to
the entity in the knowledge graph (i.e., entity in-degree)
as a simple proxy. We further break ties by favoring entities
that have a mapping to Wikipedia, and hence are ``popular''.
Note that these heuristics for breaking scoring ties are based on the
structure of the knowledge graph, as neither of these signals are
available from the surface lexical forms of the entities.

\section{Experimental Setup}

We conducted evaluations on the \textsc{SimpleQuestions}
dataset~\cite{Bordes_etal_2015}, comprised of 75.9k/10.8k/21.7k
training/validation/test questions. Each
question is associated with a (subject, predicate,
object) triple from a Freebase subset that answers the question. The subject is
given as an MID, but the dataset does not identify the entity in
the question, which is needed for our formulation of entity detection.
For this, we used the names file
by \citet{DaiZihang_etal_ACL2016} to backproject the entity names onto
the questions to annotate each token as either \textsc{Entity} or
\textsc{NotEntity}. This introduces some noise, as in some cases there
are no exact matches---for these, we back off to fuzzy matching
and project the entity onto the $n$-gram sequence with the
smallest Levenshtein Distance to the entity name. As with previous work, we
report results over the 2M-subset of Freebase.

For entity detection, we evaluate by extracting every sequence of
contiguous \textsc{Entity} tags and compute precision, recall, and
F1 against the ground truth. For both entity linking and relation
prediction, we evaluate recall at $N$ (R@$N$), i.e., whether the
correct answer appears in the top $N$ results. For end-to-end
evaluation, we follow the approach of \citet{Bordes_etal_2015} and
mark a prediction as correct if both the entity and the relation
exactly match the ground truth. The main metric is accuracy,
which is equivalent to R@1.

Our models were implemented in PyTorch v0.2.0 with
CUDA 8.0 running on an NVIDIA GeForce GTX 1080 GPU.
GloVe embeddings~\cite{Pennington_etal_EMNLP2014} of size 300 served as the
input to our models. We used negative log likelihood loss to optimize
model parameters using Adam, with an initial learning rate of 0.0001. We
performed random search over hyperparameters, exploring a range that
is typical of NNs for NLP applications; the hyperparameters were
selected based on the development set. In our final model, all LSTM
and GRU hidden states sizes and MLP hidden sizes were set to 300. For
the CNNs, we used a size 300 output channel. Dropout rate for the CNNs was
0.5 and 0.3 for the RNNs.
For the CRF implementation, we used the Stanford NER
tagger~\cite{Finkel_etal_ACL2005}. For LR, we used the
scikit-learn package in Python. For Levenshtein Distance,
we used the \texttt{ratio} function in
the ``fuzzywuzzy'' Python package.
Evidence integration involves
crossing $m$ candidate entities with $r$ candidate relations,
tuned on the validation set.

\section{Results}

We begin with results on individual components. To alleviate the
effects of parameter initialization, we ran experiments with $n$
different random seeds ($n=20$ for entity detection and $n=50$ for
relation prediction). Following \citet{Reimers_Gurevych_EMNLP2017},
and due to questions about assumptions of normality, we simply report
the mean as well as the minimum and maximum scores achieved in square
brackets.

For entity detection, on the validation set, the BiLSTM (which
outperforms the BiGRU) achieves $93.1~[92.8~93.4]$ F1, compared to the
CRF at 90.2. 
Entity linking results (R@$N$) are shown in
Table~\ref{tab:linking-results} for both the BiLSTM and the CRF.
We see that entity linking using the CRF achieves comparable accuracy,
even though the CRF performs slightly worse on entity detection alone;
entity linking appears to be the bottleneck.
Error analysis shows that there is a long tail of highly-ambiguous
entities---that is, entities in the knowledge graph that have the
same label---and that even at depth 50, we are unable to identify
the correct entity (MID) more than 10\% of the time.

\begin{table}[t]
\small \centering
\begin{tabular}{llll}
\hline
\textbf{R@$N$}      & \textbf{BiLSTM} & \textbf{CRF} \\
\hline
\hline
1   & $67.8~[67.5~68.0]$  & $66.6$  \\
5   & $82.6~[82.3~82.7]$  & $81.3$  \\
20  & $88.7~[88.5~88.8]$  & $87.4$  \\
50  & $91.0~[90.8~91.1]$  & $89.8$  \\
\hline
\end{tabular}
\vspace{-0.2cm}
\caption{Results for entity linking on the validation set, given the underlying entity detection model.}
\label{tab:linking-results}
\vspace{0.2cm}
\end{table}

\begin{table}[t]
\small \centering
\begin{tabular}{llll}
\hline
\textbf{Model}      & \textbf{R@1} & \textbf{R@5} \\
\hline
\hline
BiGRU                         & $82.3~[82.0~82.5]$  & $95.9~[95.7~96.1]$  \\
CNN                           & $82.8~[82.5~82.9]$  & $95.8~[95.7~96.1]$  \\
LR (tf-idf)                   & 72.4                & 87.6            \\
LR (GloVe+rel)                & 74.7                & 92.2            \\
\citet{Ture_Jojic_EMNLP2017}  & 81.6                & -               \\
\hline
\end{tabular}
\vspace{-0.2cm}
\caption{Results for relation prediction on the validation set using different models.}
\label{tab:rel-pred-results}
\end{table}

Results of relation prediction are shown in
Table~\ref{tab:rel-pred-results} on the validation
set. \citet{Ture_Jojic_EMNLP2017} conducted the same component-level
evaluation, the results of which we report (but none else that we
could find). We are able to achieve slightly better accuracy.
Interestingly, we see that the CNN slightly outperforms the BiGRU
(which beats the BiLSTM slightly; not shown) on R@1,
but both give essentially the same results for R@5.
Compared to LR, it seems clear that for this task NNs form
a superior solution.

Finally, end-to-end results on the test set are shown in
Table~\ref{table:e2e} for various combinations of entity detection/linking and
relation prediction. We found that crossing 50 candidate entities with
five candidate relations works the best. To compute the [min,
    max] scores, we crossed 10 randomly-selected entity models with 10
  relation models. The best model combination is BiLSTM (for entity
detection/linking) and BiGRU (for relation prediction), which achieves an
accuracy of 74.9, competitive with a cluster of recent top results.
\citet{Ture_Jojic_EMNLP2017} reported a much
higher accuracy, but we have not been able to replicate their
results (and their source code does not appear to be available online). 
Setting aside that work, we are two points away from the
next-highest reported result in the literature.

\begin{table}[t]
\small \centering
\begin{tabular}{lll}
\hline
{\bf Entity} & {\bf Relation} & \textbf{Acc.} \\
\hline
\hline
BiLSTM & BiGRU          & {\bf $74.9~[74.6~75.1]$}\\
BiLSTM & CNN            & $74.7~[74.5~74.9]$ \\
BiLSTM & LR (tf-idf)    & $68.3~[68.2~68.5]$ \\
BiLSTM & LR (GloVe+rel) & $70.9~[70.8~71.1]$ \\
CRF    & BiGRU          & $73.7~[73.4~73.9]$ \\
CRF    & CNN            & $73.6~[73.4~73.7]$ \\
CRF    & LR (tf-idf)    & 67.3 \\
CRF    & LR (GloVe+rel) & 69.9 \\
\hline
\hline
\multicolumn{2}{l}{\textbf{Previous Work}}              &  \\
\hline
\hline
\multicolumn{2}{l}{\citet{Bordes_etal_2015}}     & 62.7  \\
\multicolumn{2}{l}{\citet{Golub_He_EMNLP2016}}    & 70.9  \\
\multicolumn{2}{l}{\citet{Lukovnikov_etal_WWW2017}} & 71.2  \\
\multicolumn{2}{l}{\citet{DaiZihang_etal_ACL2016}}           & 75.7  \\
\multicolumn{2}{l}{\citet{YinWengpeng_etal_2016}}        & 76.4  \\
\multicolumn{2}{l}{\citet{YuMo_etal_ACL2017}}       & 77.0  \\
\multicolumn{2}{l}{\citet{Ture_Jojic_EMNLP2017}}       & 86.8  \\
\hline
\end{tabular}
\vspace{-0.2cm}
\caption{End-to-end answer accuracy on the test set with different
  model combinations, compared to a selection of previous results
  reported in the literature.}
\label{table:e2e}
\vspace{-0.2cm}
\end{table}

Replacing the BiLSTM with the CRF for entity detection/linking
yields 73.7, which is only a 1.2 absolute decrease in end-to-end
accuracy. Replacing the BiGRU with the CNN for relation prediction
has only a tiny effect on accuracy (0.2 decrease at most).
Results show that the baselines that don't use neural networks (CRF + LR)
perform surprisingly well:\ combining LR (GloVe+rel) or LR (td-idf)
for relation prediction with CRFs for entity detection/linking
achieves 69.9 and 67.3, respectively.  Arguably, the former still
takes advantages of neural networks since it uses word embeddings, but
the latter is unequivocally a ``NN-free'' baseline. We note that this
figure is still higher than the original \citet{Bordes_etal_2015}
paper. Cast in this light, our results suggest that
neural networks have indeed contributed to real and meaningful improvements in
the state of the art according to this benchmark dataset, but that the
improvements directly attributable to neural networks are far more
modest than previous papers may have led readers to believe.

One should further keep in mind an important caveat in interpreting the
results in Table~\ref{table:e2e}:\ As
\citet{Reimers_Gurevych_EMNLP2017} have discussed, non-determinism
associated with training neural networks can yield significant
differences in accuracy. \citet{Crane_etal_TACL2018} further
demonstrated that for answer selection in question answering, a range
of mundane issues such as software versions can have a significant
impact on accuracy, and these effects can be larger than incremental
improvements reported in the literature. We adopt the emerging
best practice of reporting results from multiple trials, but this
makes comparison to previous single-point results difficult. 

It is worth emphasizing that all NN models we have examined can
be characterized as ``Deep Learning 101'':\ easily within the grasp of
a student after taking an intro NLP course. Yet, our strong baselines
compare favorably with the state of the art. It seems that some
recent models exhibit unnecessary
complexity, in that they perform {\it worse} than our baseline.
State-of-the-art NN architectures only improve upon
our strong baselines modestly, and at the cost of introducing
significant complexity---i.e., they are ``doing a lot'' for only
limited gain. In real-world deployments, there are advantages to
running simpler models even if they may perform slightly
worse. \citet{Sculley_etal_2014} warned that machine-learned
solutions have a tendency to incur heavy technical debt in terms
of ongoing maintenance costs at the systems level. The fact that
Netflix decided not to deploy the winner of the Netflix Prize (a
complex ensemble of many different models)~is a real-world example.

\section{Conclusions}

Moving forward, we are interested in more formally characterizing
complexity--accuracy tradeoffs and their relation to the amount of
training data necessary to learn a model. It is perhaps self-evident
that our baseline CNNs and RNNs are ``less complex'' than other recent
models described in the literature, but how can we compare model
complexity objectively in a general way? The number of model
parameters provides only a rough measure, and does not capture the
fact that particular arrangements of architectural elements make
certain linguistic regularities much easier to learn. We seek to gain
a better understanding of these tradeoffs. One concrete empirical
approach is to reintroduce additional NN architectural elements in a
controlled manner to isolate their contributions. With a strong
baseline to build on, we believe that such studies can be executed
with sufficient rigor to yield clear generalizations.

To conclude, we offer the NLP community three points of
reflection:\ First, at least for the task of simple QA over knowledge
graphs, in our rush to explore ever sophisticated deep learning
techniques, we have not adequately examined simple, strong baselines
in a rigorous manner. Second, it is important to consider baselines
that do not involve neural networks, even though it is easy to forget
that NLP existed before deep learning. Our experimental results show
that, yes, deep learning is exciting and has certainly advanced the
state of the art, but the actual improvements are far more modest than
the literature suggests. Finally, in our collective frenzy to improve
results on standard benchmarks, we may sometimes forget that the
ultimate goal of science is knowledge, not owning the top entry in a
leaderboard.

\section{Acknowledgments}

This research was supported by the Natural Sciences and Engineering
Research Council (NSERC) of Canada.

\bibliography{kbqa-baseline}
\bibliographystyle{acl_natbib}

\end{document}